\documentclass{article}

\usepackage{microtype}
\usepackage{graphicx}
\usepackage{subfigure}
\usepackage{booktabs} 
\usepackage{float}
\usepackage{multirow}
\usepackage{siunitx}
\usepackage{hyperref}

\usepackage[accepted]{icml2024}

\usepackage{amsmath}
\usepackage{amssymb}
\usepackage{mathtools}
\usepackage{amsthm}
\usepackage{xcolor}
\usepackage{soul}

\usepackage[capitalize,noabbrev]{cleveref}

\theoremstyle{plain}

\theoremstyle{definition}

\theoremstyle{remark}

\usepackage[textsize=tiny]{todonotes}

\begin{document}

\twocolumn[
\icmltitle{Is persona enough for personality? Using ChatGPT to reconstruct an agent's latent personality from simple descriptions}



\icmlsetsymbol{equal}{*}

\begin{icmlauthorlist}
\icmlauthor{Yongyi Ji}{usc}
\icmlauthor{Zhisheng Tang}{isi}
\icmlauthor{Mayank Kejriwal}{isi}
\end{icmlauthorlist}

\icmlaffiliation{usc}{Department of Industrial \& Systems Engineering, University of Southern California, California, USA}
\icmlaffiliation{isi}{Information Sciences Institute, USC Viterbi School of Engineering,
4676 Admiralty Way 1001, Marina Del Rey, California 90292}

\icmlcorrespondingauthor{Mayank Kejriwal}{kejriwla@isi.edu}
\icmlcorrespondingauthor{Zhisheng Tang}{zhisheng@isi.edu}

\icmlkeywords{Machine Learning, ICML}

\vskip 0.3in
]



\printAffiliationsAndNotice{}  

\begin{abstract}
Personality, a fundamental aspect of human cognition, contains a range of traits that influence behaviors, thoughts, and emotions. This paper explores the capabilities of large language models (LLMs) in reconstructing these complex cognitive attributes based only on simple descriptions containing socio-demographic and personality type information. Utilizing the HEXACO personality framework, our study examines the consistency of LLMs in recovering and predicting underlying (latent) personality dimensions from simple descriptions. Our experiments reveal a significant degree of consistency in personality reconstruction, although some inconsistencies and biases, such as a tendency to default to positive traits in the absence of explicit information, are also observed. Additionally, socio-demographic factors like age and number of children were found to influence the reconstructed personality dimensions. These findings have implications for building sophisticated agent-based simulacra using LLMs and highlight the need for further research on robust personality generation in LLMs.
\end{abstract}

\section{Introduction}

Large language models (LLMs), such as GPT-3.5 and GPT-4, have opened up new avenues in exploring novel applications across domains spanning education \cite{gan2023large}, healthcare \cite{singhal2023large}, creative writing \cite{gomez2023confederacy}, and computational social science \cite{ziems2024can}. A growing body of work seeks to understand emergent cognitive abilities in LLMs \cite{binz2023using}, including theory of mind \cite{kosinski2023evaluating}, numeracy \cite{imani2023mathprompter}, and common sense reasoning \cite{huang2022towards}. In this paper, we consider one such cognitive aspect of LLMs; namely, to what extent can these models accurately represent and reconstruct a complex human personality type without the type being \emph{explicitly} described to the model?

Previous research has explored using classic human personality assessments, such as HEXACO \cite{lee2004psychometric} and MBTI \cite{boyle1995myers}, to evaluate the personality of LLMs \citet{miotto2022gpt,pan2023llms}. Their research has established that LLMs (at least those evaluated at the time) do possess underlying personality types. However, we are less interested in deducing the personality of the LLM itself, than in determining whether a commercial LLM like GPT-3.5 can accurately reconstruct and represent a multi-dimensional personality type based solely on simple descriptions. Personality reconstruction using LLMs is a fascinating yet challenging task, and one not sufficiently explored, to the best of our knowledge. Human personality involves many traits, behaviors, and cognitive patterns that are intertwined, often influenced by contextual factors and life experiences in ways that are still being researched \cite{hopwood2011genetic}. 

In this paper, we investigate the LLMs' ability to reconstruct an agent's latent personality type, which consists of the six dimensions detailed in Table \ref{tab:my_label} according to the HEXACO model, from simple descriptions. We also aim to evaluate whether socio-demographic descriptions guide personality reconstruction. We explore the feasibility of using LLMs for personality reconstruction and the key factors influencing the models' ability to do so. We design an extensive set of prompts and conduct a series of experiments to examine how well GPT-3.5 and GPT-4 reproduce the traits of expected personality dimensions in the HEXACO model. 

Our experiments show that LLMs are capable of reconstructing latent personality dimensions even when given simple persona descriptions. However, several inconsistencies are also observed. We also find a weak, but biased, inclination toward hallucination: the LLM reconstructs personality dimensions not used as a latent variable for prompt construction, and the dimensions are biased toward one extreme of the dimension. We also find that socio-demographic information, such as age and the number of children, significantly influences the dimensions of reconstructed personality types. However, the evidence in the real world of socio-demographic correlations with personality types is not high, which suggests that LLMs may have to be properly calibrated if we don't want socio-demographics to play a strong role in determining personality types.  

\section{Persona Construction}

To construct a single persona description, we consider both socio-demographic and personality types. For the socio-demographic, five aspects are considered: age, gender, marital status, annual household income, and number of children. The specific values associated with each of the five aspects are enumerated as follows: \textbf{Age}: This is an integer value. Four numerical intervals are defined: [18, 30), [30, 50), [50, 65), [65, 80]. Once a specific interval is chosen, an integer value will be randomly sampled from a uniform distribution of the chosen interval. \textbf{Gender}: \{male, female\}. \textbf{Marital status}: \{single, married, divorced\}. \textbf{Annual household income}: Similar to the Age aspect, this is also an integer value, and three intervals are defined: [26,500, 52,000), [52,000, 156,000), [156,000, 223,000]. Once a specific interval is chosen, an integer value will be randomly sampled from a uniform distribution of the chosen interval. \textbf{Number of children}: \{no child, one child, more than one child\}.

Additionally, we define constraints such that any person with a reasonable background would be unlikely to violate them. The three constraints are defined as follows: (1) A person who is single cannot have children. (2) A person under 19 years old cannot have children. (3) A person under 30 years old cannot be divorced.

To construct a complete socio-demographic description, we randomly sample a value from each of the five aspects described above while following the constraints. Then, the sampled values are inserted into a template: You are \textbf{Age} years old, a \textbf{Gender} who is \textbf{Marital Status} and has \textbf{Number of Children}, and your annual household income is \$\textbf{Household Income}. The bold texts in the template are places where the value corresponding to each aspect should be inserted. Notice that for the \textbf{Age} and \textbf{Annual household income} aspects, after the interval is sampled, the specific numerical number will be randomly sampled given the interval. For example, if the sampled \textbf{Age} aspect is [18, 30) and the subsequently sampled integer value is 22, the sampled \textbf{Gender} aspect is female, the sampled \textbf{Marital status} aspect is single, the sampled \textbf{Annual household income} aspect is [52,000, 156,000) and the subsequently sampled integer value is 67,000, and the sampled \textbf{Number of children} aspect is no child, the completed socio-demographic description will be: \textit{You are 22 years old, a female who is single and have no child, and your annual household income is \$67000.}


For the personality type description, we utilize the HEXACO personality test \cite{lee2004psychometric}, which measures six different dimensions of personality: Honesty-Humility, Emotionality, Extraversion, Agreeableness, Conscientiousness, and Openness to Experience. For each dimension, official descriptions that depict people with high and low scores are provided. Based on these descriptions, we use GPT-3.5-Turbo to construct four sentences for each dimension: two describing the behavior of individuals who score high on that dimension and another two for those who score low. These sentences are documented in Table \ref{tab:my_label}. The prompt used can be found in Appendix Table \ref{tab:app_two_sen}.

\begin{table*}[]
\centering
\scriptsize
\begin{tabular}{|c|p{6cm}|p{6cm}|}
\hline
Personality Dimension&  High Score& Low Score\\
\hline
Honesty-Humility&  You prefer to avoid manipulating others for personal gain and follow rules diligently. You are uninterested in seeking lavish wealth or elevated social status.& You may often manipulate others for personal gain, breaking rules without hesitation. Material wealth and self-importance drive your actions.\\
\hline
Emotionality&  You seek emotional support from others in times of stress. You do not fear physical harm and prefer emotional detachment from others.& You tend to be unfazed by physical harm and stress, and prefer to keep your concerns to yourself. You feel emotionally detached from others and worry very little.\\
\hline
Extraversion&  You enjoy social gatherings and interactions, feeling confident and positive about yourself. In contrast, you may feel awkward when you are the center of social attention, preferring to be more reserved.& You prefer quiet activities alone, such as reading or hobbies. You may feel uncomfortable in large social gatherings.\\
\hline
Agreeableness&  You often forgive and cooperate with others, able to control your temper. You tend to hold grudges and be critical, feeling anger easily at mistreatment.& You may find yourself holding onto grudges and being critical of others' flaws. You might also tend to defend your opinions stubbornly and react with anger when mistreated.\\
\hline
Conscientiousness&  You prioritize organization and accuracy in your tasks, striving for perfection in your work. You tend to deliberate carefully when making decisions, avoiding impulsivity and reflecting on your choices.& You may often prefer to avoid challenging tasks and be content with work that has some mistakes. You might make decisions impulsively without much reflection.\\
\hline
Openness to Experience&  You enjoy exploring art, nature, and new knowledge. You often ponder imaginative ideas and appreciate uniqueness.& You prefer familiar and traditional activities in your daily life. You tend to avoid engaging in creative or unconventional ideas.\\
\hline
\end{tabular}
\caption{The list of two sentences used as personality type descriptions for either a high or low score on each of the six personality dimensions.}
\label{tab:my_label}
\end{table*}

\par\nobreak

To generate a complete personality type description, we randomly select five of the six dimensions. Each chosen dimension is randomly assigned either a high or low score. Using the selected configuration, we get ten sentences describing a person's personality with the chosen dimensions of personality. Notice that we intentionally omit one personality dimension, aiming to understand the behavior of LLM when lacking the description for one dimension.

Combining one sentence of the socio-demographic description and ten sentences of the personality type description, we have a total of 11 sentences as a complete persona description. Additionally, to enhance the coherence and flow of the description, we reorder these sentences by utilizing GPT-3.5-Turbo. The prompt is detailed in Appendix Table \ref{tab:app_reoder}. An example of a final persona description is as follows: \textit{You are uninterested in seeking lavish wealth or elevated social status. You prefer quiet activities alone, such as reading or hobbies. You may often prefer to avoid challenging tasks and be content with work that has some mistakes. You seek emotional support from others in times of stress. You might make decisions impulsively without much reflection. You prefer to avoid manipulating others for personal gain and follow rules diligently. You may feel uncomfortable in large social gatherings. You might also tend to defend your opinions stubbornly and react with anger when mistreated. You do not fear physical harm and prefer emotional detachment from others. You may find yourself holding onto grudges and being critical of others' flaws. You are 22 years old, a female who is single and have no child, and your annual household income is \$67000.}

\section{Experiments}

To evaluate whether persona description is enough for LLMs to reconstruct an agent's latent personality, we apply the HEXACO personality test to the LLM with such a description. We use the 60-statement HEXACO test \cite{ashton2009hexaco}. An example of such a statement is: \textit{I would be quite bored by a visit to an art gallery.} The test taker is required to give a number between 1 and 5 that indicates how much the test taker agrees or disagrees with the statement, with 5 being strongly agree and 1 being strongly disagree. The full list of statements is detailed in Appendix Table \ref{tab:app_statement}. 

Given a complete persona description consisting of one sentence of socio-demographic description and ten sentences of personality type description, the LLM is prompted to give a score between 1 and 5, representing how much it disagrees or agrees with a given statement. The prompt used is: \textit{You will be provided with a statement about you. Please read it and decide how much you agree or disagree with that statement on the basis of your personality description. Write your response using the following scale:\textbackslash{}n\textbackslash{}n5 = strongly agree\textbackslash{}n4 = agree\textbackslash{}n3 = neutral\textbackslash{}n2 = disagree\textbackslash{}n1 = strongly disagree.\textbackslash{}n\textbackslash{}nPlease answer the statement, even if you are not completely sure of your response. The answer should be a numerical value and limited to the range of 1, 2, 3, 4, or 5, without any punctuation marks.\textbackslash{}n\textbackslash{}n\textbackslash{}n Your personality description: [Persona Description].} The \textit{[Persona Description]} is the placeholder for the given persona description. Note that the prompt described above is used as the system message, whereas the actual statement is used as the user message. 

After answering all 60 statements, we can procedurally determine if such LLM, under the influence of the given persona description, scores high or low in each of the six personality dimensions. This procedure is detailed in Appendix Section \ref{app_pro}. We aim to understand how each aspect of the socio-demographic description and each dimension of the personality type description affect the LLM's score on all six dimensions of personality. 

We randomly sampled 1000 personas and used all of them to test GPT-3.5-Turbo. Furthermore, to see if a model with a larger scale and better ability can do better, we use one-tenth of the 1000 personas (100 personas) to test GPT-4-Turbo. All experiments and persona construction use the official OpenAI API and a temperature setting of 0 to ensure reproducibility.

\section{Result}


\begin{table*}[h!]\centering
\resizebox{\textwidth}{!}{
\begin{tabular}{|c|cccccc|c|}
\hline
Dependent Variable $\rightarrow$ & \multirow{2}{*}{Honesty-Humility} & \multirow{2}{*}{Emotionality} & \multirow{2}{*}{Extraversion} & \multirow{2}{*}{Agreeableness} & \multirow{2}{*}{Conscientiousness} & \multirow{2}{*}{Openness to Experience} & \multirow{2}{*}{Aggregated} \\
Independent Variable $\downarrow$ &&&&&&& \\
\hline
Marital Status &$0.30916^{*}$ &$0.4335$ &$0.0913$ &$0.00357^{**}$ &$0.228837$ &$0.879$ &$0.64175$ \\
Age &$0.03009$ &$0.0317^{*}$ &$0.5149$ &$0.26578$ &$0.887307$ &$0.933$ &$0.00842^{**}$ \\
Annual Household Income &$0.70499$ &$9.39\num{e-12}^{***}$ &$0.2726$ &$0.50784$ &$0.488875$ &$0.533$ &$0.31571$ \\
Number of Children &$2.34\num{e-07}^{***}$ &$0.217$ &$0.9998$ &$0.01494^{*}$ &$0.395229$ &$0.393$ &$9.96e-06^{***}$ \\
Gender &$0.098$ &$8.15\num{e-05}^{***}$ &$0.5705$ &$0.18758$ &$0.000791^{***}$ &$0.163$ &$0.52233$ \\
\hline
Honesty-Humility & $<$ $2\num{e-16}^{***}$ &$<$ $2\num{e-16}^{***}$ &$0.1306$ &$1.85\num{e-10}^{***}$ &$6.51\num{e-08}^{***}$ &$7.33\num{e-06}^{***}$ &$<$ $2\num{e-16}^{***}$ \\
Emotionality &$1.93\num{e-07}^{***}$ &$<$ $2\num{e-16}^{***}$ &$0.097$ &$3.26\num{e-07}^{***}$ &$0.000358^{***}$ &$3.18\num{e-06}^{***}$ &$<$ $2\num{e-16}^{***}$ \\
Extraversion &$0.00679^{**}$ &$0.4543$ &$0.0713$ &$<$ $2\num{e-16}^{***}$ &$4.39\num{e-09}^{***}$ &$1.53\num{e-09}^{***}$ &$0.53621$ \\
Agreeableness &$1.18\num{e-06}^{***}$ &$2.91\num{e-05}^{***}$ &$0.506$ &$<$ $2\num{e-16}^{***}$ &$0.000695^{***}$ &$0.589$ &$<$ $2\num{e-16}^{***}$ \\
Conscientiousness &$3.93\num{e-15}^{***}$ &$0.653$ &$0.3361$ &$0.82573$ &$<$ $2\num{e-16}^{***}$ &$5.52\num{e-05}^{***}$ &$<$ $2\num{e-16}^{***}$ \\
Openness to Experience &$<$ $2\num{e-16}^{***}$ &$0.0318^{*}$ &$0.2175$ &$1.68\num{e-05}^{***}$ &$1.35\num{e-15}^{***}$ &$3.83\num{e-11}^{***}$ &$<$ $2\num{e-16}^{***}$\\
\hline
\end{tabular}
}
\caption{The one-way Analysis of Variance (ANOVA) tests p-values conducted using the results of the HEXACO personality test finished by GPT-3.5-Turbo prompted with 1000 personas. The row index shows the independent variable and consists of aspects and dimensions that are provided to GPT-3.5-Turbo. The column index shows the dependent variable and consists of reconstructed personality dimensions. The `Aggregated' column shows the results when the dependent variables are concatenated. *, **, and *** represent statistical significance at the 95\%, 99\%, and 99.9\% confidence levels, respectively.}\label{tab:anova}
\end{table*}

\begin{figure}[]
\centering
\includegraphics[width=0.55\columnwidth]{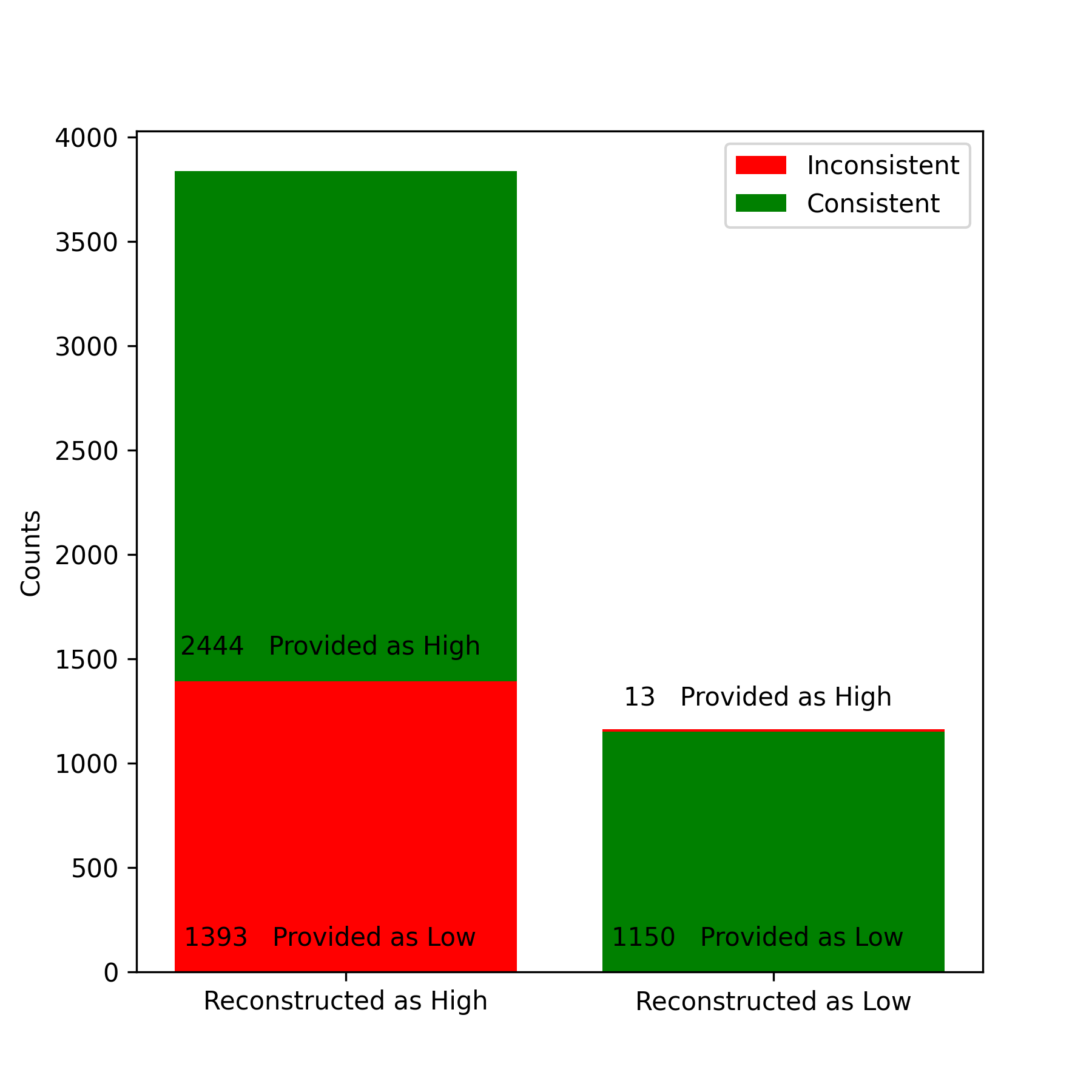}
\caption{The results of the HEXACO personality test given by personas reconstructed using GPT-3.5-Turbo provided with 1000 personality descriptions. The two columns on the left show the number of dimensions that GPT-3.5-Turbo reconstructed as high, and the two columns on the right show the number of dimensions that GPT-3.5-Turbo reconstructed as low. The two columns on the top show the number of dimensions that are provided as high in the persona descriptions given to GPT-3.5-Turbo, whereas the two columns on the bottom show the number of dimensions that are provided as low in the persona descriptions given to GPT-3.5-Turbo. Hence, the green columns represent consistency, and the red columns represent inconsistency. The results for the six individual dimensions of personality are shown in Appendix Table \ref{tab:app_overall}}\label{fig:overall}
\end{figure}

Out of 1000 personas tested, GPT-3.5-Turbo showed high consistency (71.88\%, 3594 out of 5000 dimensions) in maintaining the specified high and low scores across various personality dimensions. However, when there is a discrepancy between the provided score and the reconstructed score of a personality dimension, it is often the case (99.07\% of the time) that the LLM reconstructed a high score to a dimension that was originally provided as a low score. The Appendix Figure \ref{fig:overall} shows the consistency between the provided personality type descriptions and the reconstructed personality by GPT-3.5-Turbo and provides a more visual representation of the results. A similar trend is observed when the test is administrated on GPT-4 provided with the 100 personas descriptions. The detailed results are provided in Appendix Table \ref{tab:app_gpt4}.

\begin{figure}[]
\centering
\includegraphics[width=0.8\columnwidth]{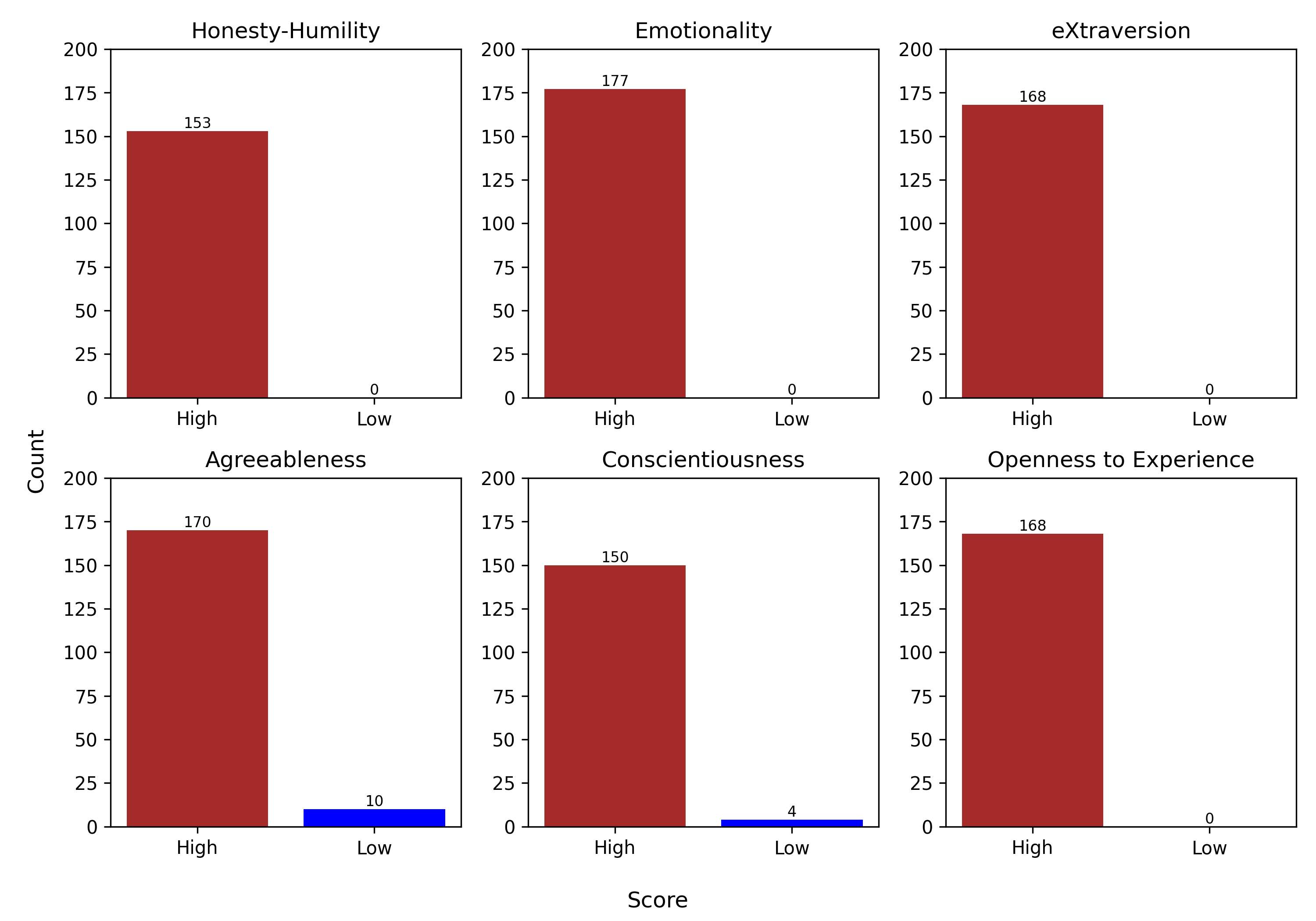}
\caption{The reconstructed scores (high or low) of each of the six personality dimensions when GPT-3.5-Turbo is provided with a personality type description that omits one of the six personality dimensions. The title of each sub-graph indicates which dimension is omitted. The result for GPT-4-Turbo is provided in Appendix Figure \ref{fig:app_omit_gpt4}.}\label{fig:omit}
\end{figure}

Through analyzing the reconstructed personality dimensions that are omitted in the persona description provided to LLMs, we find that GPT-3.5-Turbo tends to give a high score on omitted dimensions, reflecting the model's tendency to fill in missing personality dimensions with a high score. Appendix Figure \ref{fig:omit} provided a more visual presentation of the analysis and shows the score (high or low) of each of the six personality dimensions when the model is provided with personality type descriptions omitting the one dimension. A similar trend is also observed for GPT-4-Turbo.

To further understand the significance of each socio-demographic aspect and personality type dimension on the reconstructed personality type, one-way Analysis of Variance (ANOVA) tests were conducted. These tests compared the influence of each individual aspect and dimension on the six reconstructed personality dimensions, both in isolation and in aggregation as a complete personality type. Table \ref{tab:anova} presents the p-values associated with each ANOVA test. The results indicate varying levels of statistical significance across different dimensions. In general, the provided socio-demographic descriptions have an insignificant influence on three of the reconstructed personality dimensions (i.e., Extraversion, Conscientiousness, Openness to Experience) and some effect on the other three dimensions. However, almost all provided personality dimensions have a strong and significant influence on all the reconstructed dimensions, except for the Extraversion dimension. Furthermore, when we aggregate all six reconstructed personality dimensions and consider them as one personality type, we observe that the age and the number of children have a significant effect on the reconstructed personality type. Additionally, all provided personality dimensions (except for extraversion) can significantly influence the reconstructed personality type.

\section{Discussion}
Our study explores the capability of LLMs, specifically GPT-3.5 and GPT-4, to reconstruct and represent human personality types using simple persona descriptions that contain both socio-demographic and personality type information. Our experiments revealed that both models can accurately reconstruct the specified dimensions of personality type when provided with basic persona descriptions. However, mistakes are observed where the models reconstructed unintended high scores for certain dimensions. Additionally, when a personality dimension was omitted, the models tended to assign high scores to that specific dimension, reflecting a definite bias. This highlights the models' ability to reproduce given dimensions reliably but also their struggle with the unspecified aspects of personality.

The ANOVA test analysis reveals that socio-demographic information had varying levels of influence on the reconstructed personality dimensions, with age and the number of children being the significant factors. This finding points to the importance of including comprehensive socio-demographic information in persona descriptions. Additionally, detailed personality descriptions were found to be critical in guiding LLMs to accurately reconstruct specific personality dimensions, with significant influence on almost all reconstructed dimensions, except for the Extraversion dimension. Altogether, these results highlight the critical role of both socio-demographic and personality type information in shaping the models' predictions and outputs.

The ability of LLMs to reconstruct human-like personalities based on simple descriptions opens new avenues for building sophisticated agent-based simulacra. However, the observed biases and inconsistencies in personality reconstruction suggest the need for further research and evaluation of these LLMs to ensure accurate representations of diverse human personalities. Future research can focus on exploring methods to mitigate biases and finding more robust personality generation techniques. Our research contributes to the broader understanding of LLMs' cognitive capabilities and limitations.







\section*{Impact Statement}
LLMs are increasingly being used in computational social science, and in constructing realistic `agents' that can be studied and evaluated. Personality is an important component of any such agent that is expected to be human-like. This work adds to the emerging science of LLM-based agent construction.

\bibliography{example_paper}
\bibliographystyle{icml2024}


\newpage
\appendix
\onecolumn
\section{Appendix}

\begin{table}[]
\centering
\begin{tabular}{|p{14cm}|}
\hline
The prompt for constructing the two sentence descriptions \\
\hline
Based on personality description, generate two separate sentences about what you tend to do in daily life. Express in a simple way. Each sentence needs to be similar in length. Every sentence needs to end with a full stop.\\
\hline
\end{tabular}\caption{The prompt used to construct the two sentence descriptions for either a high score or a low score on each personality dimension. This prompt is used as the system prompt, and the official description of the specific personality dimension is used as the user message.}\label{tab:app_two_sen}
\end{table}

\begin{table}[]
\centering
\begin{tabular}{|p{14cm}|}
\hline
The prompt for reordering the 11 sentences of personality descriptions \\
\hline
You are given multiple sentences. Without modifying or adding or omitting any of the original sentences, you need to randomly put these sentences together into a single paragraph. Do not omit any original sentences. The output paragraph must contain the exact same number of sentences as the given sentences.\\
\hline
\end{tabular}\caption{The prompt used to reorder the 11 sentences of personality descriptions. This prompt is used as the system prompt, and the 11 sentences of personality descriptions are used as the user message.}\label{tab:app_reoder}
\end{table}

\begin{table}
    \centering
    \scriptsize
    \begin{tabular}{|c|p{4cm}|c|p{4cm}|c|p{4cm}|} \hline 
         Index&  Statement&  Index&  Statement&  Index& Statement\\ \hline 
         1&  I would be quite bored by a visit to an art gallery.&  21& People think of me as someone who has a quick temper.  &  41& I can handle difficult situations without needing emotional support from anyone else.\\ \hline 
         2&  I plan ahead and organize things, to avoid scrambling at the last minute.&  22&  On most days, I feel cheerful and optimistic. &  42& I would get a lot of pleasure from owning expensive luxury goods.\\ \hline 
         3&  I rarely hold a grudge, even against people who have badly wronged me.&  23&  I feel like crying when I see other people crying. &  43& I like people who have unconventional views.\\ \hline 
         4&  I feel reasonably satisfied with myself overall.&  24&  I think that I am entitled to more respect than the average person is. &  44& I make a lot of mistakes because I don’t think before I act.\\ \hline 
         5&  I would feel afraid if I had to travel in bad weather conditions.&  25&  If I had the opportunity, I would like to attend a classical music concert. &  45& Most people tend to get angry more quickly than I do.\\ \hline 
         6&  I wouldn't use flattery to get a raise or promotion at work, even if I thought it would succeed.&  26&  When working, I sometimes have difficulties due to being disorganized. &  46& Most people are more upbeat and dynamic than I generally am.\\ \hline 
         7&  I'm interested in learning about the history and politics of other countries.&  27&  My attitude toward people who have treated me badly is “forgive and forget”. &  47& I feel strong emotions when someone close to me is going away for a long time.\\ \hline 
         8&  I often push myself very hard when trying to achieve a goal.&  28&  I feel that I am an unpopular person. &  48& I want people to know that I am an important person of high status.\\ \hline 
         9&  People sometimes tell me that I am too critical of others.&  29&  When it comes to physical danger, I am very fearful. &  49& I don’t think of myself as the artistic or creative type.\\ \hline 
         10&  I rarely express my opinions in group meetings. &  30&  If I want something from someone, I will laugh at that person's worst jokes. &  50& People often call me a perfectionist.\\ \hline 
         11&  I sometimes can't help worrying about little things. &  31&  I’ve never really enjoyed looking through an encyclopedia. &  51& Even when people make a lot of mistakes, I rarely say anything negative.\\ \hline 
         12&  If I knew that I could never get caught, I would be willing to steal a million dollars. &  32&  I do only the minimum amount of work needed to get by.  &  52& I sometimes feel that I am a worthless person.\\ \hline 
         13&  I would enjoy creating a work of art, such as a novel, a song, or a painting. &  33&  I tend to be lenient in judging other people. &  53& Even in an emergency I wouldn’t feel like panicking.\\ \hline 
         14&  When working on something, I don't pay much attention to small details .&  34& In social situations, I’m usually the one who makes the first move.  &  54& I wouldn’t pretend to like someone just to get that person to do favors for me.\\ \hline 
         15&  People sometimes tell me that I'm too stubborn .&  35&  I worry a lot less than most people do. &  55& I find it boring to discuss philosophy.\\ \hline 
         16&  I prefer jobs that involve active social interaction to those that involve working alone .&  36&  I would never accept a bribe, even if it were very large. &  56& I prefer to do whatever comes to mind, rather than stick to a plan.\\ \hline 
         17&  When I suffer from a painful experience, I need someone to make me feel comfortable .&  37&  People have often told me that I have a good imagination. &  57& When people tell me that I’m wrong, my first reaction is to argue with them.\\ \hline 
         18&  Having a lot of money is not especially important to me .&  38&  I always try to be accurate in my work, even at the expense of time. &  58& When I’m in a group of people, I’m often the one who speaks on behalf of the group.\\ \hline 
         19&  I think that paying attention to radical ideas is a waste of time .&  39&  I am usually quite flexible in my opinions when people disagree with me. &  59& I remain unemotional even in situations where most people get very sentimental.\\ \hline 
         20&  I make decisions based on the feeling of the moment rather than on careful thought .&  40&  The first thing that I always do in a new place is to make friends. &  60& I’d be tempted to use counterfeit money, if I were sure I could get away with it.\\ \hline
    \end{tabular}
    \caption{The complete 60 statements of the HEXACO test.}
    \label{tab:app_statement}
\end{table}

\begin{table*}[]
\centering
\resizebox{\textwidth}{!}{%
\begin{tabular}{|c|cc|cc|cc|}
\hline
&\multicolumn{2}{c|}{\textbf{Honesty-Humility}} &\multicolumn{2}{c|}{\textbf{Emotionality}} &\multicolumn{2}{c|}{\textbf{Extraversion}} \\
&Reconstructed as High &Reconstructed as Low &Reconstructed as High &Reconstructed as Low &Reconstructed as High &Reconstructed as Low \\
Provided as High &426 &0 &410 &0 &396 &0 \\
Provided as Low &182 &239 &18 &395 &433 &3 \\
\hline
&\multicolumn{2}{c|}{\textbf{Agreeableness}} &\multicolumn{2}{c|}{\textbf{Conscientiousness}} &\multicolumn{2}{c|}{\textbf{Openness to Experience}} \\
&Reconstructed as High &Reconstructed as Low &Reconstructed as High &Reconstructed as Low &Reconstructed as High &Reconstructed as Low \\
Provided as High &399 &13 &420 &0 &393 &0 \\
Provided as Low &29 &379 &331 &95 &400 &39 \\
\hline
\end{tabular}%
}
\caption{The results of the HEXACO personality test given by personas reconstructed using GPT-3.5-Turbo provided with 1000 personality descriptions, shown separately for each of the six dimensions of personality.}\label{tab:app_overall}
\end{table*}

\begin{table*}[]
\centering
\resizebox{\textwidth}{!}{%
\begin{tabular}{|c|cc|cc|cc|}
\hline
&\multicolumn{2}{c|}{\textbf{Honesty-Humility}} &\multicolumn{2}{c|}{\textbf{Emotionality}} &\multicolumn{2}{c|}{\textbf{Extraversion}} \\
&Reconstructed as High &Reconstructed as Low &Reconstructed as High &Reconstructed as Low &Reconstructed as High &Reconstructed as Low \\
Provided as High &43 &0 &34 &0 &36 &0 \\
Provided as Low &6 &33 &0 &50 &16 &25\\
\hline
&\multicolumn{2}{c|}{\textbf{Agreeableness}} &\multicolumn{2}{c|}{\textbf{Conscientiousness}} &\multicolumn{2}{c|}{\textbf{Openness to Experience}} \\
&Reconstructed as High &Reconstructed as Low &Reconstructed as High &Reconstructed as Low &Reconstructed as High &Reconstructed as Low \\
Provided as High &19 &32 &46 &0 &41 &0 \\
Provided as Low &0 &30 &40 &3 &25 &21 \\
\hline
\end{tabular}%
}
\caption{The results of the HEXACO personality test given by personas reconstructed using GPT-4-Turbo provided with 100 personality descriptions, shown separately for each of the six dimensions of personality.}\label{tab:app_gpt4}
\end{table*}

\begin{figure}[]
\centering
\includegraphics[width=0.8\columnwidth]{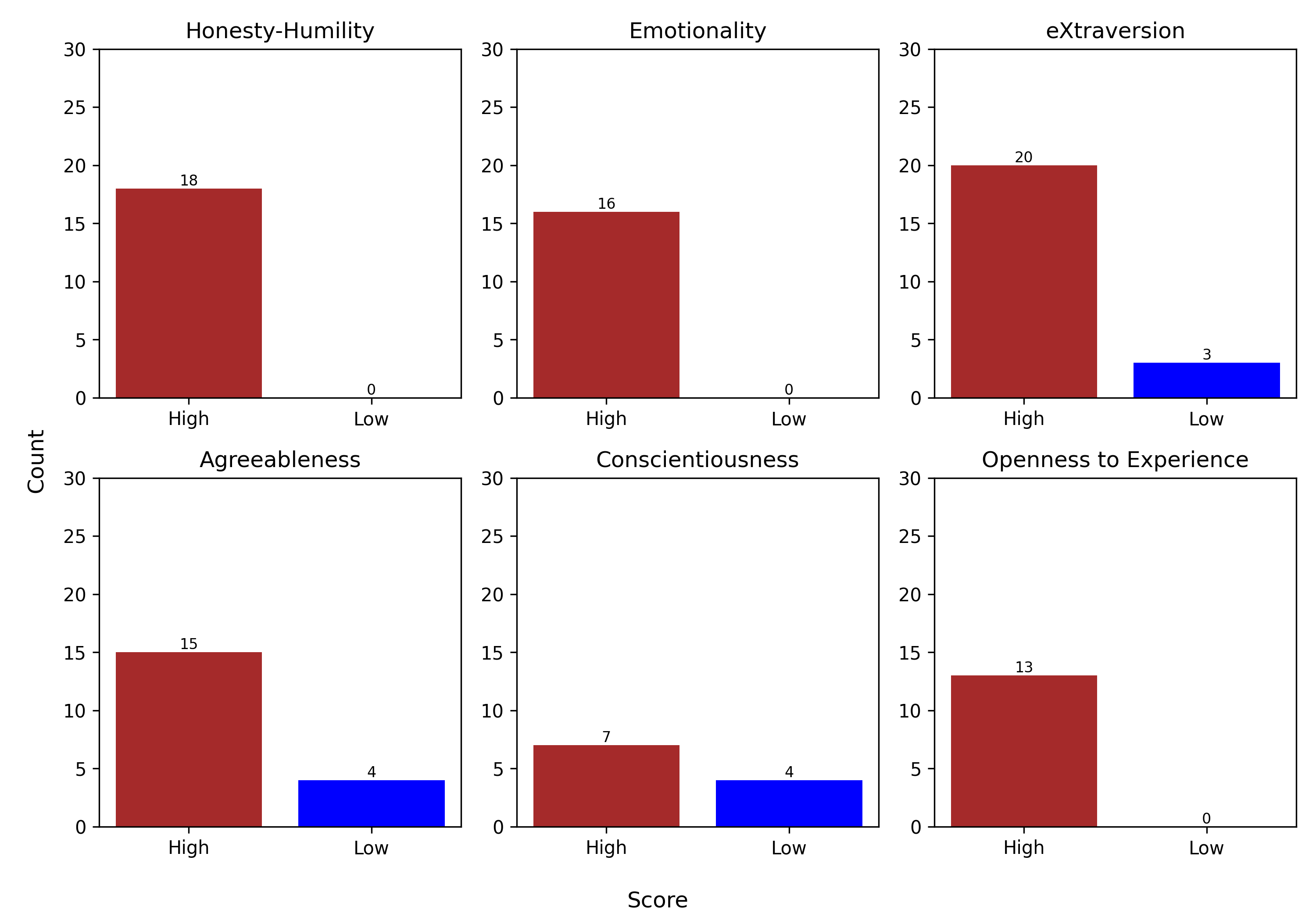}
\caption{The reconstructed scores (high or low) of each of the six personality dimensions when GPT-4-Turbo is provided with a personality type description that omits one of the six personality dimensions. The title of each sub-graph indicates which dimension is omitted.}\label{fig:app_omit_gpt4}
\end{figure}

\subsection{Transformation Procedure}\label{app_pro}

According to the official HEXACO personality test instructions, we need to map the raw scores from certain statements to their reverse values. The mapping rule is detailed in Table \ref{tab:mapping_score}. The \textbf{R} symbol that follows immediately after the statement index indicates that the score from this specific statement needs to be reversed. Scores of 5 are mapped to 1, 4 to 2, 3 remains unchanged, 2 becomes 4, and 1 becomes 5. After mapping the scores, we calculate the scores for each personality dimension by averaging the scores of the relevant indices associated with each dimension, which is also indicated in Table \ref{tab:mapping_score}. Finally, we determine the direction of each personality dimension by comparing the associated score with 2.5. Scores larger than 2.5 are assigned as high scores, while scores less than or equal to 2.5 are assigned as low scores.

\begin{table}
    \centering
    \scriptsize
    \begin{tabular}{|c|p{3cm}|c|p{3cm}|c|p{3cm}|} \hline 
         Dimension&  Index&  Dimension&  Index&  Dimension& Index\\ \hline 
         Honesty-Humility&  6, 30R, 54, 12R, 36, 60R, 18, 42R, 24R, 48R &  Extraversion &  4, 28R, 52R, 10R, 34, 58, 16, 40, 22, 46R &  Conscientiousness& 2, 26R, 8, 32R, 14R, 38, 50, 20R, 44R, 56R\\ \hline 
         Emotionality&  5, 29, 53R, 11, 35R, 17, 41R, 23, 47, 59R &  Agreeableness&  3, 27, 9R, 33, 51, 15R, 39, 57R, 21R, 45 &  Openness to Experience& 1R, 25, 7, 31R, 13, 37, 49R, 19R, 43, 55R\\ \hline
    \end{tabular}
    \caption{The HEXACO personality test scoring rules.}
    \label{tab:mapping_score}
\end{table}

\end{document}